\pdfoutput=1   
\documentclass[twocolumn, switch]{article}
\usepackage{preprint}
\usepackage[utf8]{inputenc}
\usepackage[T1]{fontenc}
\usepackage{libertine}
\usepackage[libertine]{newtxmath}
\usepackage{amsmath,amsfonts}
\usepackage{amssymb}
\usepackage{xcolor}
\definecolor{darkblue}{rgb}{0.1,0.2,0.6}
\usepackage{graphicx}
\usepackage{multirow}
\usepackage[colorlinks=true,
            linkcolor=black,
            urlcolor=darkblue,
            citecolor=darkblue,
            anchorcolor=black,
            hypertexnames=false]{hyperref}
\usepackage{xurl}
\usepackage{cite}
\usepackage{placeins}
\usepackage{float}
\usepackage{tikz}
\usepackage{pgfplots}
\usepackage{authblk}
\pgfplotsset{
  compat=1.17,
  every axis/.append style={
    tick label style={font=\footnotesize},
    label style={font=\footnotesize},
    legend style={font=\footnotesize},
  },
}

\graphicspath{{figures/}{./}}

\title{Low Cost Two-Stage Fabric Defect Detection at the Edge}

\author[1]{Rasel Hossen$^{*}$\thanks{Equal contribution. \texttt{hrasel2002@gmail.com}}}
\author[1]{Diptajoy Mistry$^{*}$\thanks{Equal contribution. \texttt{diptomistry50@gmail.com}}}
\author[1]{Mosaddek Hossain Kamal\thanks{\texttt{tushar@du.ac.bd}}}
\affil[1]{Department of Computer Science and Engineering, University of Dhaka, Dhaka, Bangladesh}

\begin{document}
\twocolumn[\begin{@twocolumnfalse}
\maketitle

\begin{abstract}
Fabric inspection in the garment industries of low-income economies remains largely manual, and commercial vision systems are priced beyond most small and medium mills. Because defects are sparse under controlled production, a natural response is a cascade: screen every frame with a cheap anomaly detector and invoke a full detector only on suspicious frames. We build such a cascade for four knit-fabric defect classes and deploy it end-to-end on an NVIDIA Jetson Nano with TensorRT FP16. Stage~1 is a compact convolutional autoencoder with decoder attention gates, an edge-weighted reconstruction loss, and feature-level distillation from a frozen YOLOv5n teacher; Stage~2 is YOLOv5n, invoked only on flagged frames. On a 249-image benchmark disjoint from detector training (20 defective, 229 non-defective), Stage~1 at a recall-prioritised threshold flags all 20 defective images (95\% CI 0.83--1.00) at a false-positive rate of \textbf{49.3\%} (113/229), reducing false positives by 19.3\% relative to a plain autoencoder ($p=0.011$). The parallel pipeline reaches \textbf{13.45\,FPS} against \textbf{9.86\,FPS} for a sequential YOLO-only loop. Our central finding comes from decomposing that $1.36\times$: \textbf{91\% of it is attributable to overlapping JPEG decode with inference rather than to the cascade}, which contributes only a 5.1\% inference reduction at the measured forwarding rate $p=0.534$. We further show that forwarding here is false-positive-limited rather than prevalence-limited---85\% of forwarded frames are false alarms---and quantify the 29--45\% inference reduction attainable under tighter calibration. We report this as a caution for cascade speedups measured without controlling the data path, and position the system as AI-assisted triage rather than autonomous acceptance.
\end{abstract}

\keywords{Fabric defect detection, autoencoder, anomaly detection, YOLO, edge computing, cascade}
\vspace{0.35cm}
\end{@twocolumnfalse}]

\section{Introduction}
\label{sec:introduction}

Bangladesh is the world's second-largest exporter of ready-made garments. The
sector earns over 39 billion USD annually and contributes more than 81\% of
national exports \cite{xinhua2025, newage2025, apparelsources2025}. Maintaining
that scale depends on reliable quality control, yet fabric inspection is still
largely manual: operators watch moving fabric rolls for holes, oil spots, and
other faults, and fatigue and operator-to-operator variation make the process
inconsistent and slow \cite{wiseguy2026, verifiedmarket2025}. Automated
inspection systems exist, but most are priced and engineered for settings with
larger budgets and stronger infrastructure than typical local mills can support
\cite{li2021, kahraman2023}. The constraint that matters in small and medium
factories is not detection accuracy in the abstract; it is cost, power, and
what can be made to run on a \$100-class embedded device.

Industrial textile manufacturing does not assume defect-free output. Standards
such as AQL and the ASTM 4-point system define acceptable defect thresholds
rather than enforcing zero defects~\cite{iso2859_1_2019,astm_d5430_21}, which
implies that defect occurrence is inherently sparse; even so, those small
percentages accumulate into material loss across a production shift. Sparsity
suggests an obvious systems response, and one with a long history in computer
vision \cite{viola2001rapid}: do not run a heavy detector on every frame.
Screen cheaply, reject most frames, and reserve full detection for the
suspicious minority.

We build that cascade and deploy it end to end. Stage~1 is a compact
convolutional autoencoder that scores reconstruction error against a threshold
$\tau$; Stage~2 is a YOLOv5n detector that localises and classifies defects
only on flagged frames. The deployed runtime does not serialise the two stages
per frame: a loader thread, the Stage-1 main thread, and a Stage-2 worker
process run concurrently over a bounded queue
(Section~\ref{sec:parallel_execution}). Everything is measured on an NVIDIA
Jetson Nano with TensorRT FP16 engines.

\paragraph{What we found, including what did not work.}
The cascade runs, and the parallel pipeline is measurably faster end to end
than a sequential YOLO-only loop: \textbf{13.45\,FPS} versus
\textbf{9.86\,FPS}, a $1.36\times$ ratio. But a $1.36\times$ wall-clock ratio
is not the same as a $1.36\times$ saving from cascading, and the two are easy
to conflate. When we decompose the gain into its inference and non-inference
parts using our own per-stage timings
(Section~\ref{sec:where_the_speedup_goes}), \textbf{91\% of it is attributable
to overlapping JPEG decode with inference}; the cascade itself accounts for
8.9\%, because at the measured forwarding rate $p = 0.534$ Stage~2 still runs
on more than half of all frames. On this hardware, ingest---not
inference---is the binding constraint: non-inference time is 53.63\,ms per
image for the sequential baseline, more than the 47.77\,ms spent inside
TensorRT. A cascade cannot save what is not being spent on the detector.

We treat this as a result rather than an embarrassment. Cascade speedups
reported without controlling the data path are difficult to interpret, and we
give the accounting that makes ours interpretable. We also show \emph{why}
$p=0.534$ is high: decomposing $p$ into its true-positive and false-positive
contributions (Section~\ref{sec:forwarding_analysis}) shows that 85\% of
forwarded frames are false alarms, so on this benchmark forwarding is limited
by Stage-1 \emph{specificity}, not by defect prevalence. That reframes the
engineering problem and identifies calibration, not architecture, as the
highest-leverage next step.

\paragraph{Contributions.}
\begin{enumerate}
\item A complete two-stage fabric-defect cascade deployed on a Jetson Nano
  with TensorRT FP16 and a concurrent producer--consumer runtime, described in
  enough detail to reproduce, including a Stage-1 autoencoder that reaches
  20/20 defect recall on our benchmark at 41,441 parameters
  (Sections~\ref{sec:methodology}--\ref{sec:setup}).
\item A decomposition of the measured $1.36\times$ end-to-end speedup into
  cascade and data-path components, showing the gain is dominated by decode
  overlap and that a comparably threaded YOLO-only baseline would likely erase
  it (Section~\ref{sec:where_the_speedup_goes}). We state this as a
  measurement caution for the cascade literature.
\item An analysis of what limits forwarding rate, establishing that our
  operating point is false-positive-limited rather than prevalence-limited, and
  quantifying the 29--45\% inference reduction available under tighter
  calibration (Section~\ref{sec:forwarding_analysis}).
\end{enumerate}

\noindent Section~\ref{sec:related} discusses prior work,
Section~\ref{sec:problem_statement} defines the problem,
Section~\ref{sec:methodology} presents the method,
Section~\ref{sec:setup} describes data and protocol,
Section~\ref{sec:results} reports results,
Section~\ref{sec:limitations} states limitations, and
Section~\ref{sec:conclusion} concludes.

\section{Related Work}
\label{sec:related}

Research on automated fabric defect detection has progressed from handcrafted
feature engineering to deep learning \cite{li2021, kahraman2023}. Early methods
relied on manually designed texture descriptors and classical image processing,
which struggled to generalise across weave patterns and illumination
conditions. With convolutional networks, accuracy and generalisation improved
considerably, motivating adoption in textile quality control.

\subsection{Supervised fabric defect detection}

Wei et al.\ \cite{wei2019fabric} applied Faster R-CNN to fabric defect
localisation, using a region proposal network to generate candidates and a
classification head to identify defect types; the two-stage proposal
architecture incurs computational overhead that limits throughput on
constrained hardware. Jing et al.\ \cite{jing2022fabric} proposed an improved
YOLOv3 detector, reclustering anchor boxes with $k$-means for textile defect
dimensions and reducing total error detection rate from 3.28\% to 1.76\% on a
lattice fabric dataset. Mao and Hong \cite{mao2025} reviewed YOLO architectures
from YOLOv1 to YOLOv11 for real-time textile inspection, documenting the
progressive incorporation of attention mechanisms and lightweight backbones.
Jin et al.\ \cite{app15063228} introduced backbone and attention modifications
to YOLOv8 for multi-scale defect patterns, and Sun et al.\ \cite{sun2025}
developed an improved YOLOv8n for seamless fabric inspection with SPPF\_LSKA
multi-scale extraction and CARAFE upsampling. Liu et al.\ \cite{liu2022}
explored multi-scale feature fusion for difficult defect categories.

These methods improve detection quality, but two properties limit their
relevance to low-cost deployment. First, the detector is applied to every input
frame regardless of content, which is wasteful when defects are rare under
controlled production~\cite{iso2859_1_2019,astm_d5430_21}. Second, evaluation
is typically on high-performance GPUs rather than on embedded devices under
power and memory constraints, leaving a gap between reported benchmark
performance and deployability.

\subsection{Unsupervised anomaly detection}

Anomaly detection offers a label-efficient alternative, training only on normal
samples and flagging inputs whose reconstruction error or feature distance
deviates from learned normality. Bergmann et al.\ \cite{bergmann2018improving}
showed that autoencoders trained with structural similarity loss yield more
discriminative error for defect segmentation than MSE counterparts, and
introduced the MVTec AD benchmark \cite{bergmann2019mvtec} that became the
standard reference. Pang et al.\ \cite{Pang2021} and Ruff et al.\
\cite{ruff2021unifying} survey the field, identifying reconstruction error and
feature-space distance as the dominant paradigms. PaDiM
\cite{defard2021padim} models patch-level feature distributions as multivariate
Gaussians over a pretrained backbone; PatchCore \cite{roth2022patchcore}
maintains a coreset memory bank of normal patch features and scores by
nearest-neighbour distance, reaching near-perfect AUROC on MVTec AD. More
recent work targets the accuracy--latency frontier directly: FastFlow
\cite{yu2021fastflow} uses normalising flows over backbone features, RD4AD
\cite{deng2022rd4ad} reverses the student--teacher direction to reduce
representation leakage, SimpleNet \cite{liu2023simplenet} attains strong AUROC
with a deliberately simple feature adaptor, and EfficientAD
\cite{batzner2024efficientad} reports millisecond-scale latency with a reduced
student--teacher pair. Wang et al.\ \cite{wang2022industrial} note that
reconstruction-based methods remain well suited to edge deployment owing to
their label-free training requirement.

We do not claim to advance the anomaly-detection state of the art, and we do
not benchmark against these methods here; Section~\ref{sec:limitations} states
this omission explicitly as the principal evidential gap in our Stage-1
evaluation. Our Stage~1 is chosen for parameter count and latency on a Jetson
Nano, and the contribution of this paper concerns the deployed system and its
timing behaviour rather than screening accuracy relative to modern baselines.

\subsection{Attention and distillation in autoencoders}

Attention has been used to improve reconstruction selectivity. Wang et al.\
\cite{wang2023fcaead} proposed FCAE-AD, applying attention gates in
encoder-to-decoder skip connections to suppress background reconstruction and
amplify anomalous error; the gate formulation follows the attention-gate
construction of Schlemper et al.\ \cite{schlemper2019attention}. On
distillation, Bergmann et al.\ \cite{bergmann2020} trained student ensembles to
replicate a pretrained teacher's feature maps on normal data, flagging regions
of high student--teacher discrepancy at test time. Salehi et al.\
\cite{salehi2021} distilled teacher features at multiple pyramid levels. Wu et
al.\ \cite{aekd2024_industrial} introduced AEKD, combining reconstruction with
distillation and explicitly addressing the over-generalisation risk that arises
when a student mimics the teacher too closely on anomalous inputs; Zhang et
al.\ \cite{sensor2023_feature_distillation} reported that feature-level
distillation combined with reconstruction improves separability across
industrial surface datasets. Our use of distillation \cite{hinton2015distilling}
differs from discrepancy-based scoring: we use a supervised detector as a
frozen feature teacher purely as an auxiliary training signal, while the
inference-time anomaly score remains the autoencoder's own reconstruction
error.

\subsection{Edge deployment and cascades}

Song et al.\ \cite{song2021} deployed EfficientDet-D0 for fabric defect
detection on a Jetson TX2 with TensorRT quantisation, reaching 22.7\,FPS with
response time 2.5 times lower than cloud inference---the closest prior work to
ours in hardware and application. TensorRT \cite{nvidia2019tensorrt} provides
the FP16 and INT8 optimisation required to fit detection models within embedded
power and memory budgets. These works confirm that hardware-aware optimisation
yields measurable gains, but they do not reduce the cost of running a detector
on every frame.

Cascading a cheap filter before an expensive classifier dates to Viola and
Jones \cite{viola2001rapid}, whose face detector rejects most negative windows
at early stages and runs the full classifier on a small fraction of candidates.
The principle generalises wherever the positive class is rare. Our contribution
is therefore not the cascade idea, which is well established; to our knowledge,
however, this design principle has not previously been validated end-to-end for
fabric inspection on a Jetson-class device with a full accounting of where the
resulting wall-clock time is spent. That accounting is where we depart from
prior work: as Section~\ref{sec:where_the_speedup_goes} shows, on this hardware
the data path rather than the detector dominates wall time, which means a
cascade's compute saving and its throughput gain can differ by an order of
magnitude. We are not aware of prior fabric-inspection work that separates
these two quantities, and we suggest that reported cascade speedups be
interpreted with the ingest path controlled.

\section{Problem Statement}
\label{sec:problem_statement}

High-accuracy defect detectors are too computationally heavy to run on every
frame under the real-time and power constraints of low-cost edge hardware, while
lightweight anomaly screeners can produce excessive false positives. Defect
sparsity suggests a cascade: a cheap screener filters normal frames and a
compact detector inspects only the suspicious minority.

This work asks three questions.

\begin{enumerate}
\item \textbf{Can the cascade be deployed at all on a \$100-class device?}
  Both stages must fit in 4\,GB of shared memory within a 10\,W envelope, with
  FP16 TensorRT engines, and sustain a useful frame rate.
\item \textbf{Does a cascade's compute saving translate into throughput?}
  These are distinct quantities. Compute saving depends on the forwarding rate;
  throughput depends additionally on image ingest, host copies, queueing, and
  synchronisation. We measure both and report them separately rather than
  reporting a single end-to-end ratio.
\item \textbf{What actually limits the forwarding rate?} The fraction $p$ of
  frames reaching Stage~2 depends on both defect prevalence and Stage-1
  specificity. Which term dominates determines whether the next engineering
  step is recalibration, a better screener, or neither.
\end{enumerate}

\noindent We answer these on a single 249-image benchmark disjoint from
detector training, and we are explicit throughout about the limits that a
20-defect evaluation set places on the strength of any recall claim
(Section~\ref{sec:limitations}).

\section{Methodology}
\label{sec:methodology}

\begin{figure}[H]
\centering
\includegraphics[width=\columnwidth]{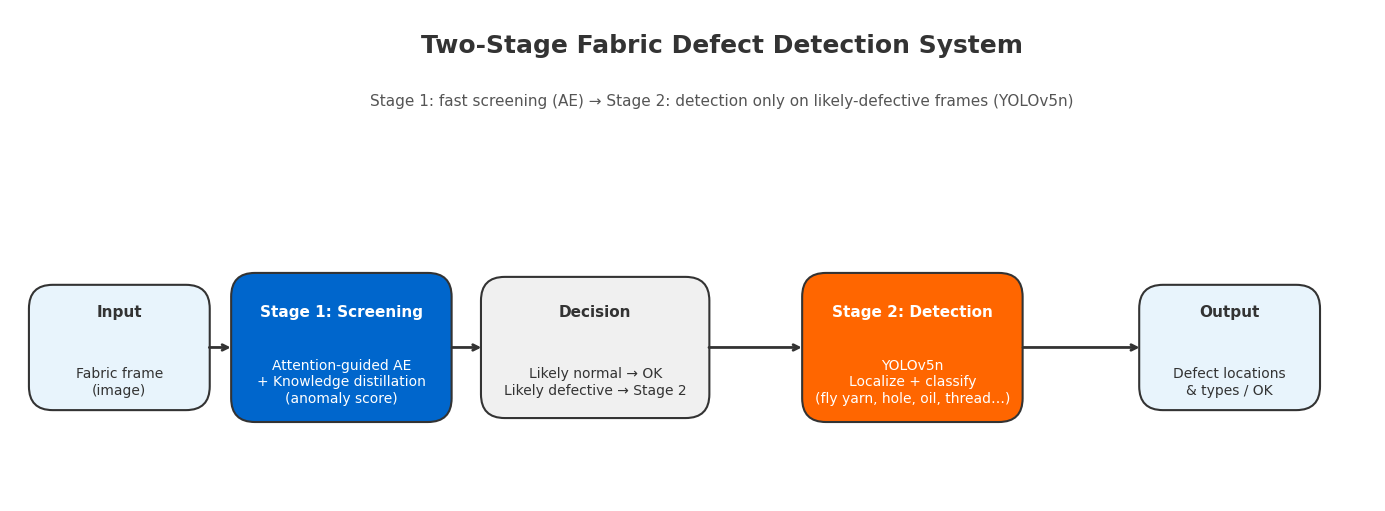}
\caption{\textbf{Two-stage fabric defect detection (logical flow).} Stage~1
filters normal frames; Stage~2 runs only on flagged anomalies. Deployment uses
a pipelined producer--consumer layout
(Section~\ref{sec:parallel_execution}), not strict AE-then-YOLO serialisation
on every frame.}
\label{fig:pipeline}
\end{figure}

\noindent\textbf{System overview.}
Fig.~\ref{fig:pipeline} shows the logical data flow. Input frames are
preprocessed and passed to Stage~1 (autoencoder), which computes a
reconstruction error and compares it against a calibrated threshold $\tau$.
Frames scoring below $\tau$ are classified as normal and discarded; frames
exceeding $\tau$ are flagged and forwarded to Stage~2 (YOLOv5n) for
localisation and classification. The figure is a logical ordering, not a
wall-clock one: the deployed runtime overlaps the stages across different
frames (Section~\ref{sec:parallel_execution}), and that distinction is
essential to interpreting our timing results.

\begin{figure}[H]
\centering
\includegraphics[width=\columnwidth]{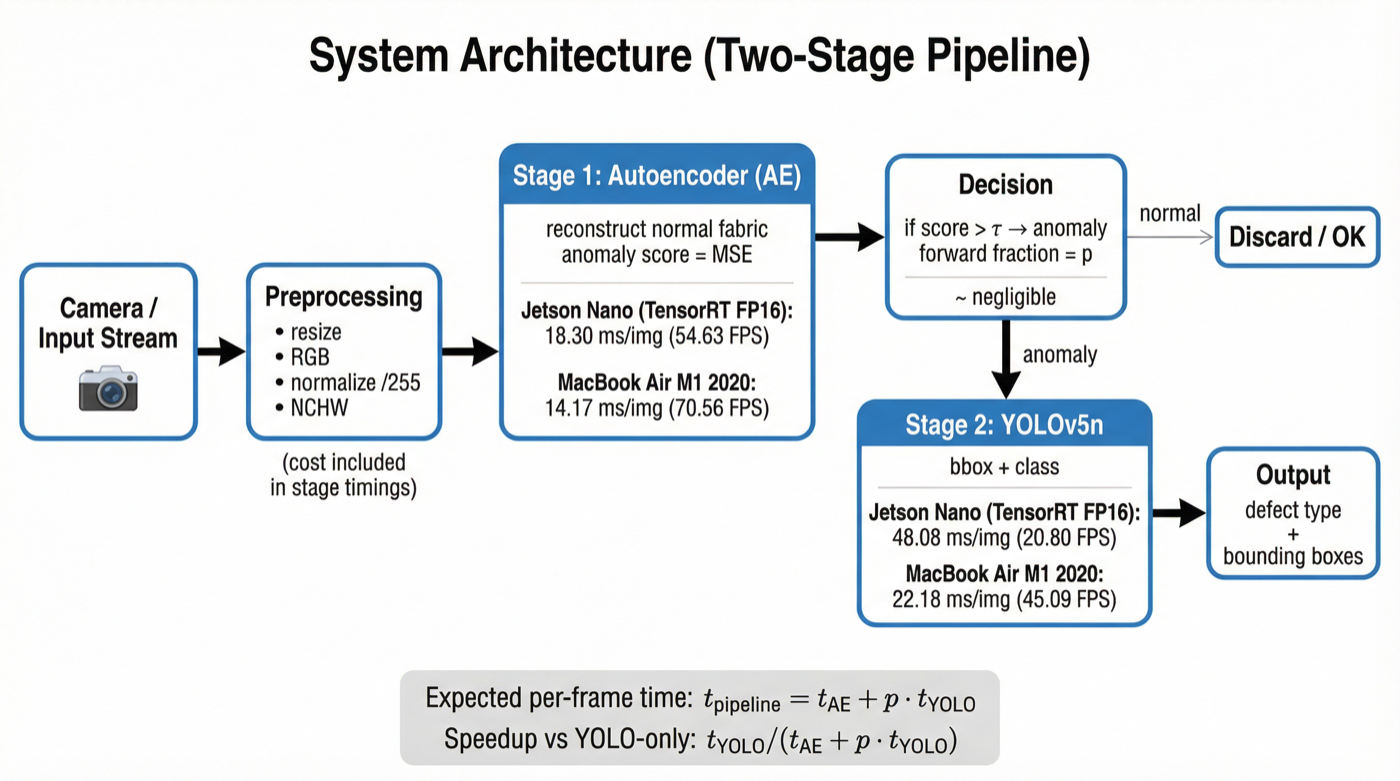}
\caption{\textbf{System architecture with per-stage latency.} Camera input is
preprocessed and screened by Stage~1 (AE). Only anomalous frames are forwarded
to Stage~2 (YOLO) for localisation and classification.}
\label{fig:system_arch_latency}
\end{figure}

\subsection{Defect Types}

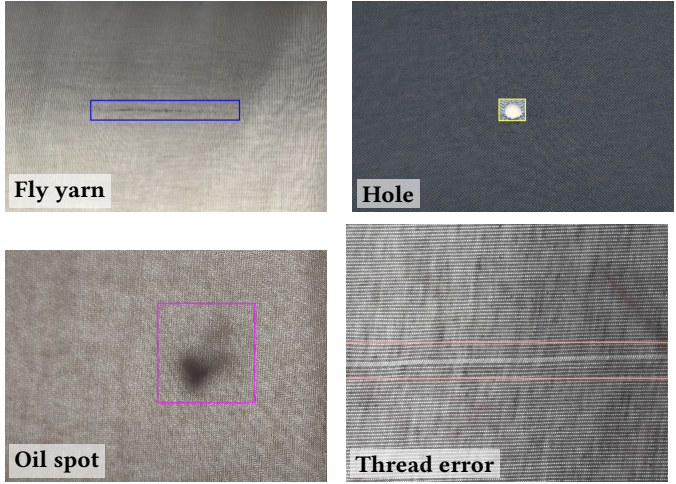
\begin{figure}[H]
\centering
\begin{tikzpicture}
  \node[anchor=south west, inner sep=0] at (0,0)
    {\includegraphics[width=0.48\columnwidth]{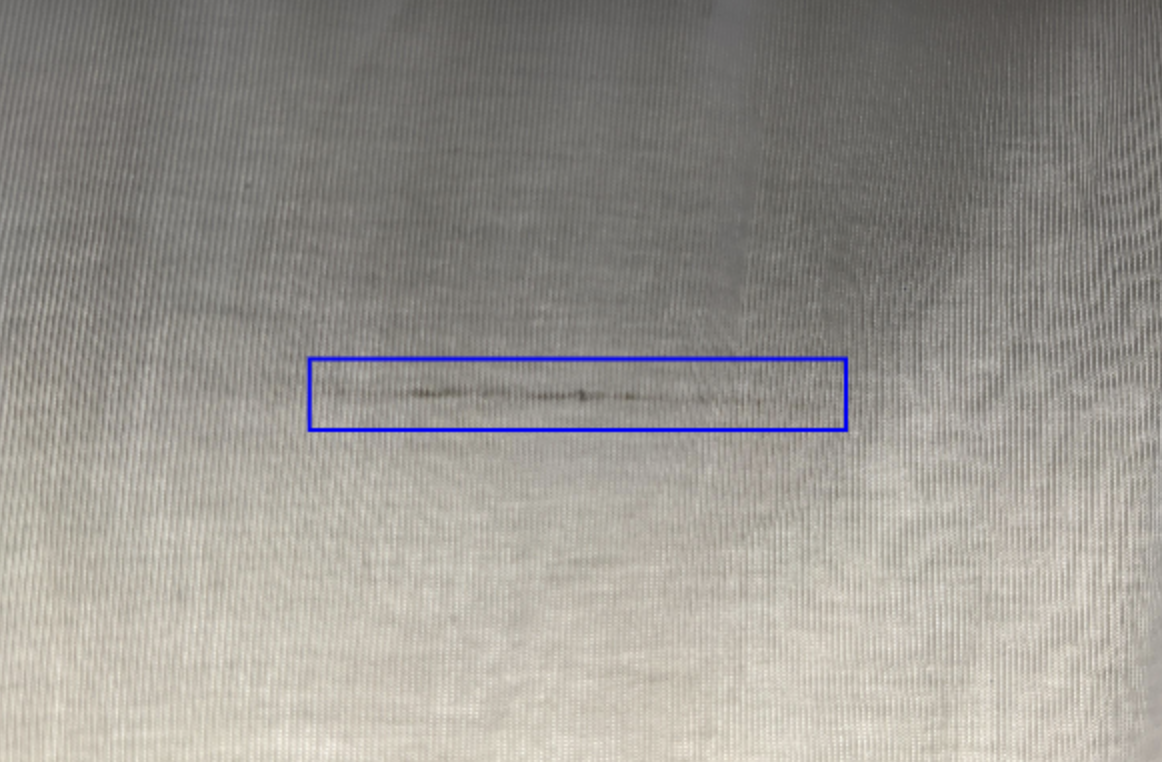}};
  \node[anchor=south west, fill=white, fill opacity=0.6,
        text opacity=1, font=\footnotesize\bfseries, inner sep=2pt] at (0.05,0.05) {Fly yarn};
\end{tikzpicture}\hfill
\begin{tikzpicture}
  \node[anchor=south west, inner sep=0] at (0,0)
    {\includegraphics[width=0.48\columnwidth]{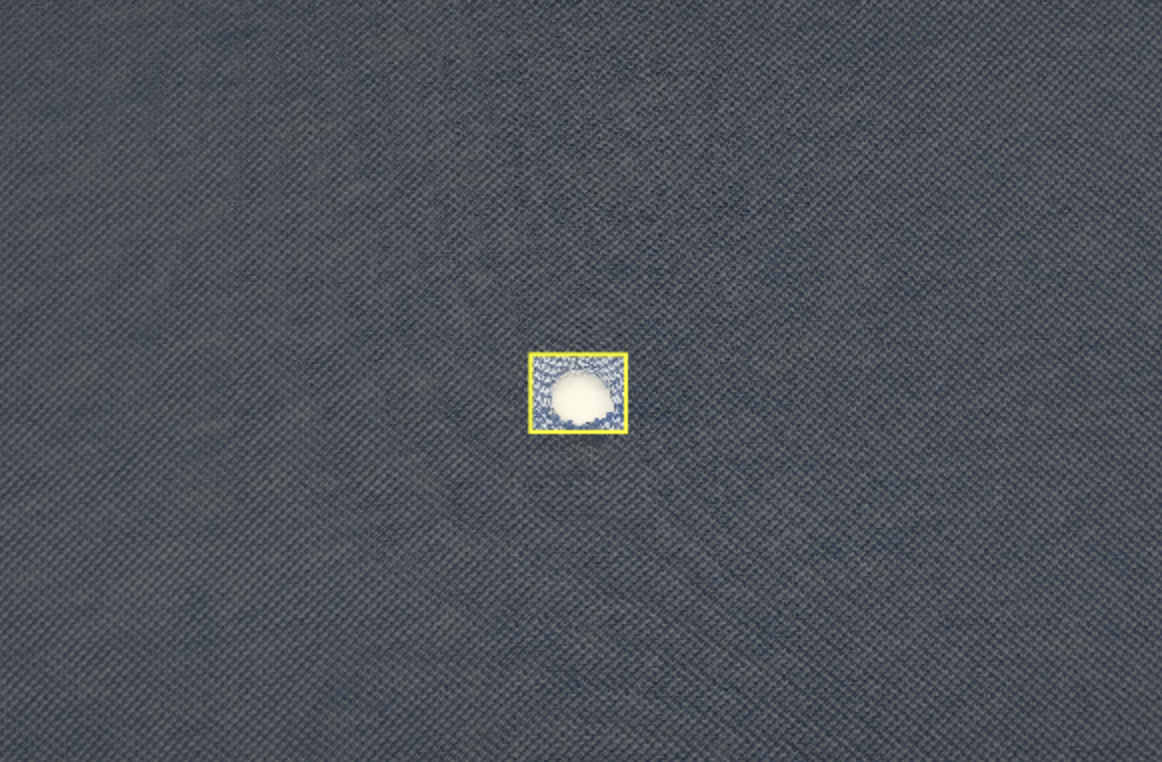}};
  \node[anchor=south west, fill=white, fill opacity=0.6,
        text opacity=1, font=\footnotesize\bfseries, inner sep=2pt] at (0.05,0.05) {Hole};
\end{tikzpicture}\\[4pt]
\begin{tikzpicture}
  \node[anchor=south west, inner sep=0] at (0,0)
    {\includegraphics[width=0.48\columnwidth]{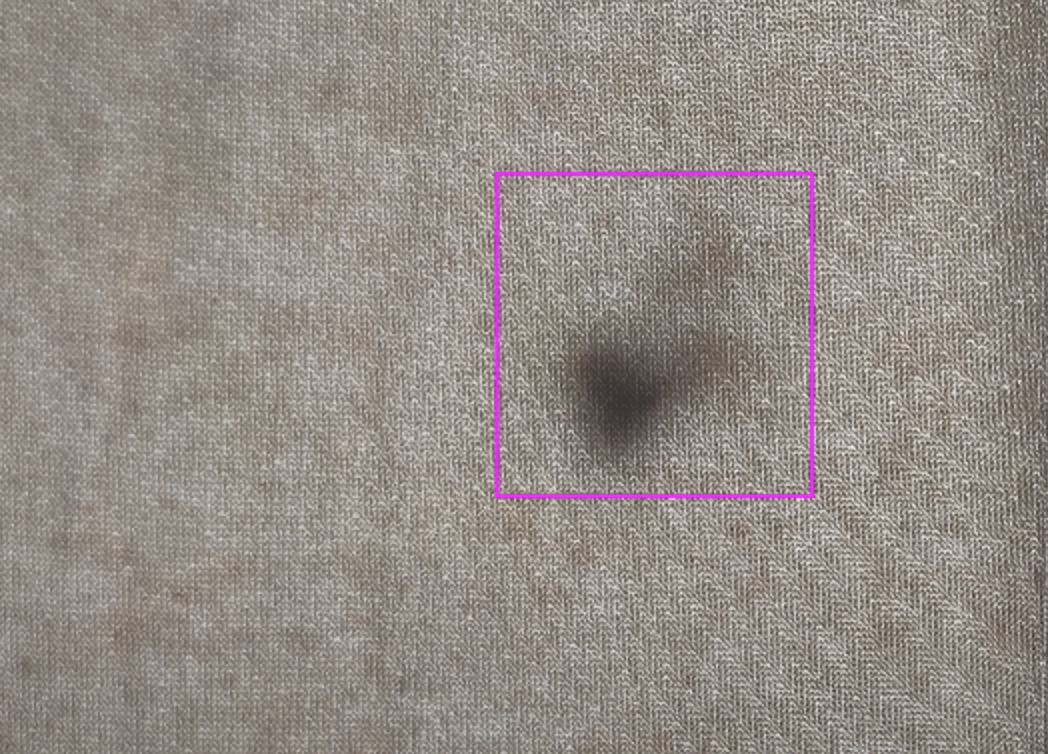}};
  \node[anchor=south west, fill=white, fill opacity=0.6,
        text opacity=1, font=\footnotesize\bfseries, inner sep=2pt] at (0.05,0.05) {Oil spot};
\end{tikzpicture}\hfill
\begin{tikzpicture}
  \node[anchor=south west, inner sep=0] at (0,0)
    {\includegraphics[width=0.48\columnwidth]{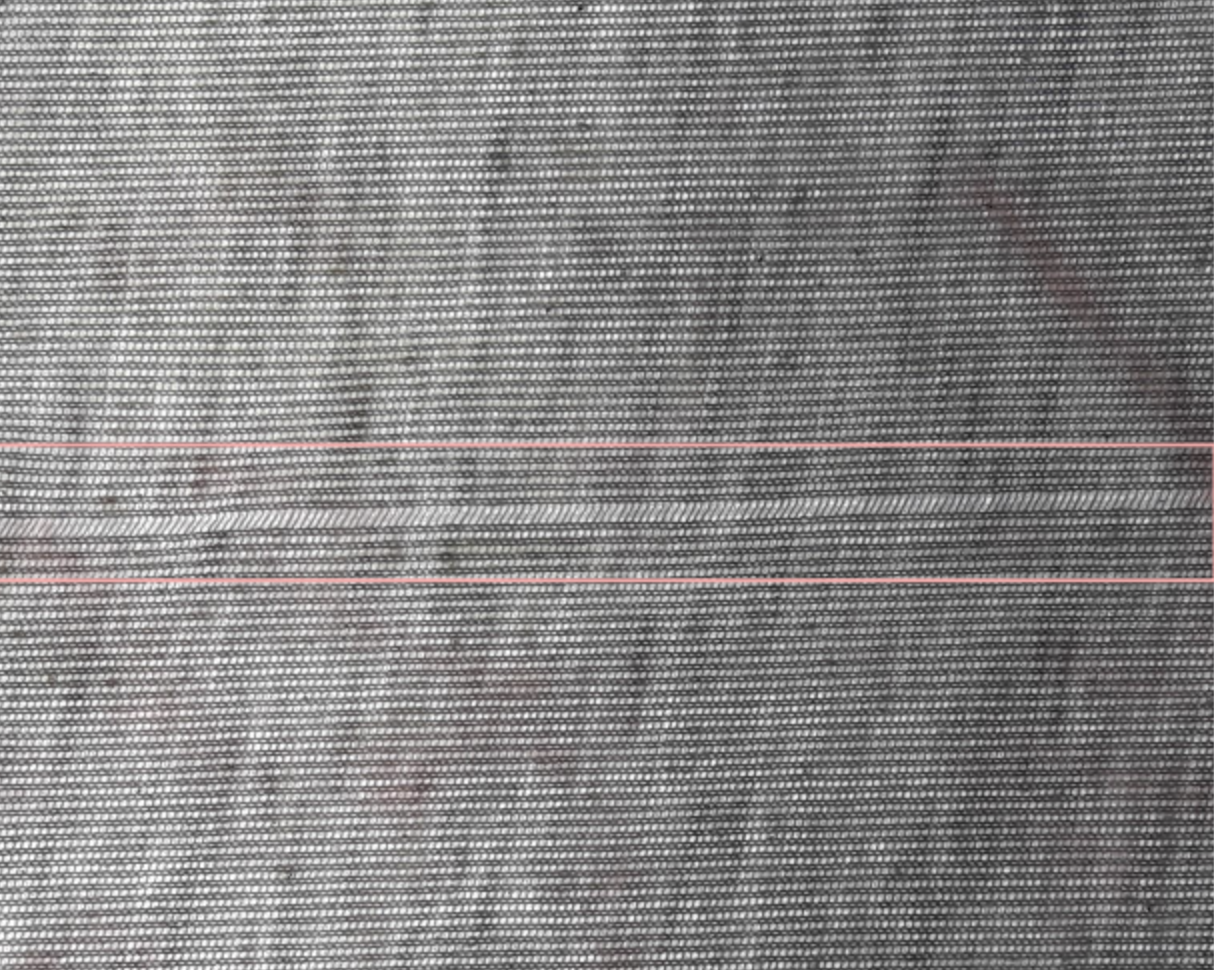}};
  \node[anchor=south west, fill=white, fill opacity=0.6,
        text opacity=1, font=\footnotesize\bfseries, inner sep=2pt] at (0.05,0.05) {Thread error};
\end{tikzpicture}
\caption{\textbf{The four defect types} with in-image labels.}
\label{fig:defect_classes}
\end{figure}

We target four defect categories common in knit production: \emph{fly yarn},
\emph{hole}, \emph{oil spot}, and \emph{thread error}
(Fig.~\ref{fig:defect_classes}). Fly yarn is loose or extraneous fibre lying
across the surface; holes are missing yarns or punctures; oil spots arise from
lubricant contamination; thread errors capture misweaves, thick places, and
misaligned yarns. These span both localised structural damage and
appearance-based contamination, and correspond to the classes annotated in our
detection dataset.

\subsection{Stage~1: Autoencoder Anomaly Screening}

Stage~1 is a lightweight, recall-prioritised filter. We train a compact
convolutional autoencoder on defect-free patches only, so that normal fabric
reconstructs with low error while unseen structure reconstructs poorly. Three
modifications are applied relative to a plain MSE autoencoder: decoder
attention gates, an edge-weighted reconstruction loss, and feature-level
distillation from a frozen YOLOv5n teacher. We refer to the plain model as the
\emph{base} autoencoder and to the model carrying all three modifications as the
\emph{modified} autoencoder. We note at the outset that these three
modifications are evaluated only as a bundle
(Section~\ref{sec:stage1_results}); we did not ablate them individually, and we
therefore make no claim about which one is responsible for the observed
improvement.

\subsubsection{Base Autoencoder}

\begin{figure}[H]
\centering
\includegraphics[width=\columnwidth]{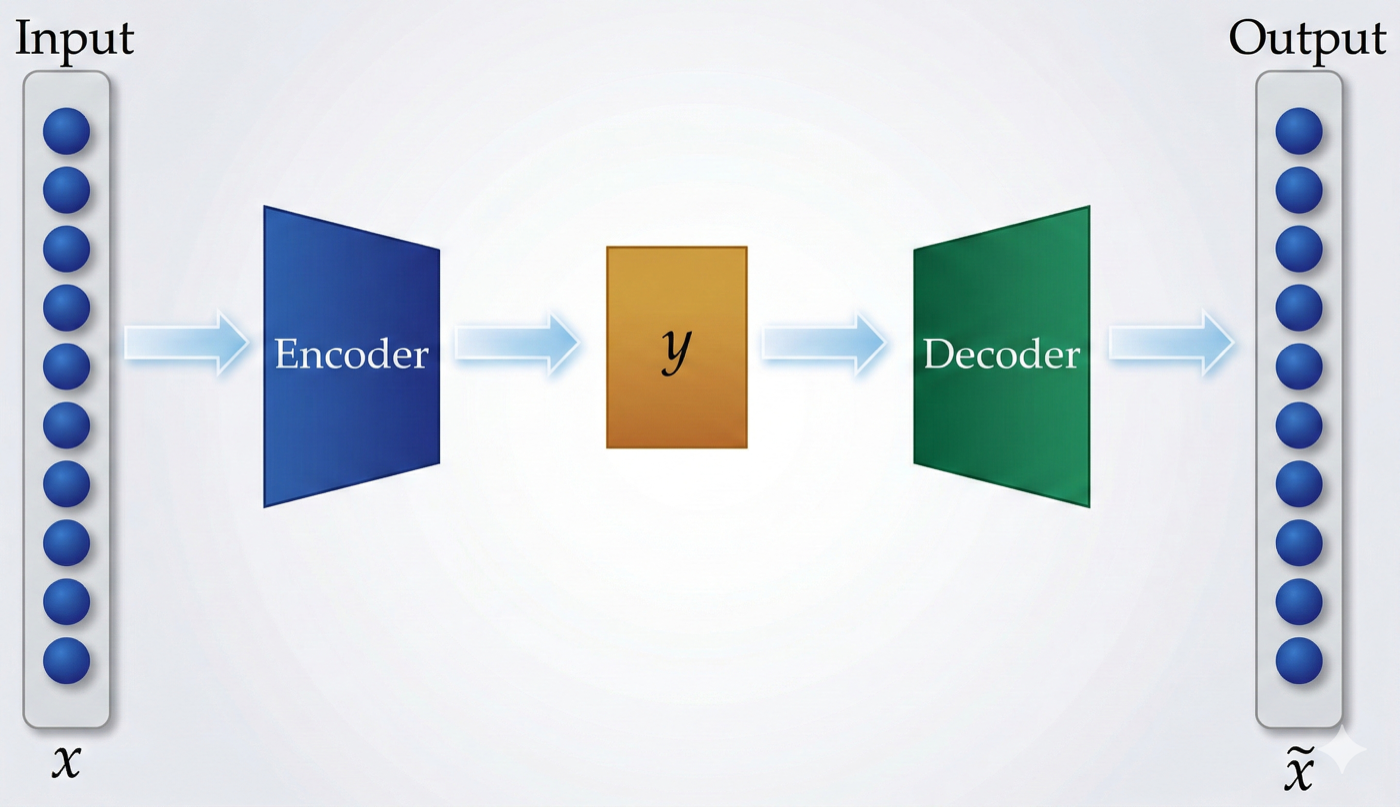}
\caption{\textbf{Base autoencoder.} Encoder compresses input $x$ to latent $y$;
decoder reconstructs $\tilde{x}$; training minimises
$\mathcal{L} = \|x - \tilde{x}\|_2$.}
\label{fig:base_ae}
\end{figure}

Our Stage~1 builds on the standard autoencoder formulation
\cite{ruff2021unifying}. An encoder maps input $x$ to a low-dimensional latent
$y = f(x, M, b)$, and a decoder reconstructs $\tilde{x} = f'(y, \tilde{M},
\tilde{b})$. Training minimises $\mathcal{L} = \|x - \tilde{x}\|^2$ so that the
latent space captures the dominant structure of the training distribution.
Trained only on normal data, the model yields higher error on inputs that
deviate from learned normality. The base configuration uses plain MSE with no
attention and no distillation (Fig.~\ref{fig:base_ae}).

\subsubsection{Architecture}

\begin{figure}[H]
\centering
\includegraphics[width=\columnwidth]{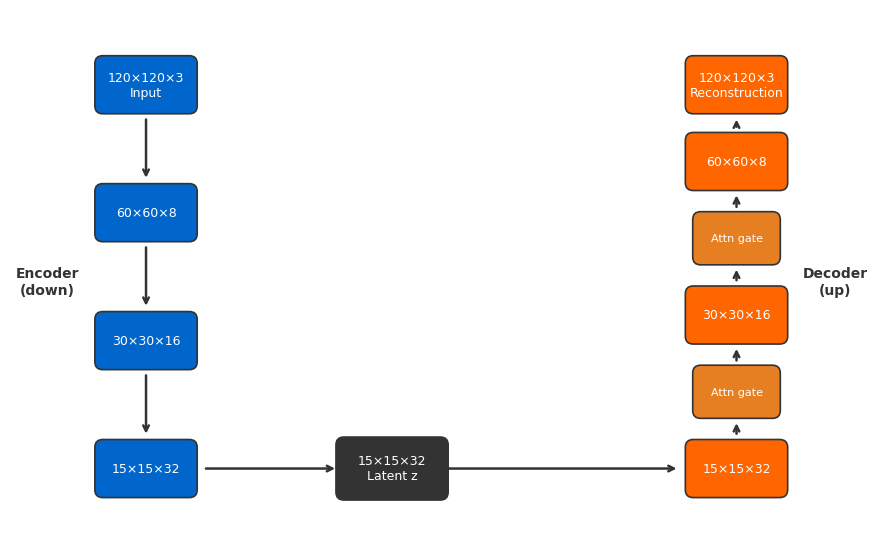}
\caption{\textbf{Modified autoencoder:} encoder, bottleneck, and decoder with
attention gates.}
\label{fig:ae_arch}
\end{figure}

The autoencoder is a symmetric encoder--decoder with \textbf{41,441} parameters
in the deployed modified configuration (\textbf{39,987} for the base
configuration without attention gates), small enough for real-time edge
inference.
The encoder accepts RGB input resized to $120\times120\times3$ and applies three
convolutional blocks, each a $3\times3$ convolution, batch normalisation, ReLU,
and stride-2 downsampling, with channel widths 8, 16, and 32. A projection step
produces a latent of size $15\times15\times48$. The decoder mirrors this
pathway with transposed convolutions. Attention gates are inserted at two
levels of the decoder, modulating features before upsampling
(Fig.~\ref{fig:ae_arch}).

\subsubsection{Attention Gates}

Attention gating has been used to improve reconstruction-based anomaly detection
by allocating reconstruction capacity preferentially to normal background
structure \cite{wang2023fcaead}. We adopt the additive attention-gate
formulation of Schlemper et al.\ \cite{schlemper2019attention}, but apply it
\emph{within} the decoder rather than across an encoder--decoder skip
connection: the gating signal $g$ is the decoder feature entering the block and
$x$ is the feature map being modulated, both drawn from the decoder path. The
attention coefficient is
\[
\alpha = \sigma\!\left( \psi\left( \mathrm{ReLU}\left( \mathrm{Conv}(g) + \mathrm{Conv}(x) \right) \right) \right),
\]
where $\mathrm{Conv}(\cdot)$ is a $1\times1$ convolution applied separately to
$g$ and $x$, $\psi$ is a $1\times1$ convolution followed by batch
normalisation, and $\sigma$ is the sigmoid. The gated feature is $\hat{x} =
\alpha \odot x$.

We are deliberately precise about what this is and is not. It is not
self-attention in the query--key sense: no pairwise similarity is computed, and
$\alpha$ is a learned spatial gate. Furthermore, when $g$ and $x$ are the same
tensor the two $1\times1$ branches are linearly redundant and collapse to a
single convolution; we retain the two-branch form because $g$ and $x$ differ in
our decoder blocks, and we note the degenerate case explicitly so that the
formulation is not misread as a general-purpose attention mechanism.

\subsubsection{Edge-Weighted Reconstruction Loss}

To bias the model toward edge fidelity we use an edge-weighted MSE:
\begin{equation}
\mathcal{L}_{\text{rec}} = \frac{1}{N} \sum_{i=1}^{N} W(i) \left( I_i - \hat{I}_i \right)^2,
\label{eq:edge_loss}
\end{equation}
where $W(i) = 1 + \lambda \cdot E(i)$ upweights pixels with strong edge
content, and $E(i)$ is the normalised Sobel gradient magnitude on the luminance
channel,
\[
E(i) = \frac{\| \nabla I(i) \|}{\max_j \| \nabla I(j) \|},
\]
with $\lambda = 0.5$ selected by a small validation sweep. The gradient becomes
$\partial \mathcal{L}_{\text{rec}} / \partial \hat{I}_i = W(i)\cdot 2(\hat{I}_i
- I_i)$, so edge pixels receive a larger learning signal.

\paragraph{Why this may cut both ways.}
The intended effect is that benign edge-like structure (weave, shadows, minor
creases) reconstructs well and yields low error, while defects introducing novel
edge structure yield high error. However, the same emphasis that improves
reconstruction of normal edges can also improve reconstruction of
\emph{defective} edges, suppressing exactly the error signal Stage~1 depends on.
This is the over-generalisation failure mode identified for
distillation-augmented autoencoders by Wu et al.\
\cite{aekd2024_industrial}, and it applies to edge weighting as well. We
therefore treat the direction of this effect as an empirical question rather
than a property established by the mechanism, and we report the bundled result
in Section~\ref{sec:stage1_results} without attributing it to edge weighting
specifically.

\subsubsection{Distillation from a Frozen YOLO Teacher}

\begin{figure}[H]
\centering
\includegraphics[width=\columnwidth]{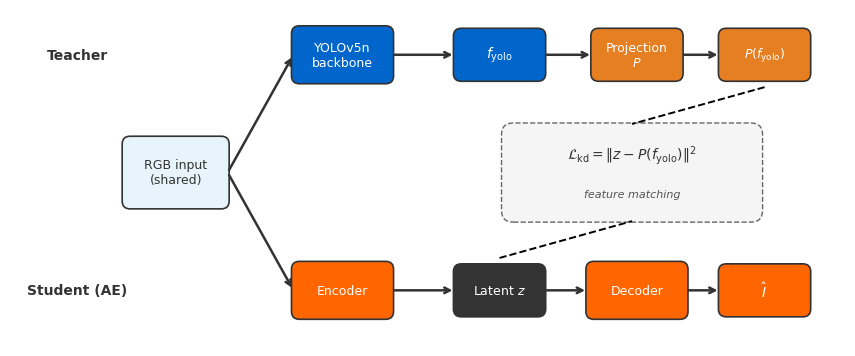}
\caption{Distillation framework. A frozen YOLOv5n teacher provides
defect-aware features, projected to match the autoencoder's latent space. The
autoencoder aligns its latent representation with these features while
minimising reconstruction error. At inference, only the autoencoder (left
branch) runs.}
\label{fig:knowledge_distill}
\end{figure}

An autoencoder trained only on normal data has no signal telling it what a
defect is. We therefore add a feature-matching term that transfers
defect-relevant representation from a supervised detector
\cite{hinton2015distilling}. The teacher is a YOLOv5n detector trained on the
labelled fabric dataset (Section~\ref{sec:setup}) and frozen thereafter. Given
input $I$, we extract teacher features $f_{\text{yolo}}$ from the CSPDarknet
backbone's last stage, prior to the detection head, where the representation is
spatially condensed but still discriminative. A lightweight projection head $P$
(adaptive average pooling followed by a $1\times1$ convolution) maps them into
the autoencoder latent space so that $P(f_{\text{yolo}})$ matches $z \in
\mathbb{R}^{15\times15\times48}$ (Fig.~\ref{fig:knowledge_distill}).

The training objective is
\begin{equation}
\mathcal{L}_{\text{total}} = \mathcal{L}_{\text{rec}} + \lambda_{\text{kd}} \, \mathcal{L}_{\text{kd}},
\qquad
\mathcal{L}_{\text{kd}} = \| z - P(f_{\text{yolo}}) \|_2^2,
\label{eq:total_loss}
\end{equation}
with $\mathcal{L}_{\text{rec}}$ from Eq.~\eqref{eq:edge_loss}. We swept
$\lambda_{\text{kd}} \in \{0.0, 0.1, 0.3, 0.5\}$ and used $0.3$; larger values
degraded reconstruction on normal fabric. Our formulation differs from
student--teacher discrepancy scoring \cite{bergmann2020, salehi2021}: the
teacher shapes the latent space during training only, and the inference-time
anomaly score remains the autoencoder's own reconstruction error.

Gradients propagate through the autoencoder and projection head only; the
teacher stays frozen, which also lets one pretrained detector be reused across
autoencoder runs. Training the teacher first and freezing it gave more stable
convergence than joint fine-tuning in our runs.

\paragraph{Inference cost.}
The teacher runs only during training. At inference the autoencoder runs alone,
so distillation adds no deployment cost beyond the autoencoder's own
parameters. This is a hard constraint rather than a convenience: the Jetson
Nano's 4\,GB of shared memory and 10\,W envelope cannot host a concurrent YOLO
backbone for screening. Our measured Stage-1 latency is 22.67\,ms per frame in
the deployed pipeline (Section~\ref{sec:jetson_results}), which is what the
pipeline sustains rather than a budget we set in advance.

\subsubsection{Thresholding}
\label{sec:thresholding}

Stage~1 requires a scalar threshold $\tau$ on mean reconstruction error. We
describe two values, and we are explicit about which one produced the reported
results and how it was chosen, because this bears directly on how the recall
figure should be read.

\paragraph{Reference threshold.}
After training we compute reconstruction errors on a held-out set of normal-only
samples and take the 99th percentile, giving $\tau_{\text{ref}} \approx 0.0189$
(stored with the exported model as \texttt{anomaly\_stats.json}). By
construction this admits roughly 1\% of normal samples, which on 229
non-defective benchmark images would be about 2 false positives.

\paragraph{Threshold actually used, and how it was selected.}
All results in this paper use $\tau = 0.00245$, which is $7.7\times$
\emph{lower} than $\tau_{\text{ref}}$. Lowering $\tau$ enlarges the flagged set
and is conservative for recall. This value was chosen on the 249-image
benchmark to prioritise defect recall, and at this setting Stage~1 flags all 20
defective images. \textbf{The threshold was therefore selected on the same
images on which recall is reported}, so the 20/20 figure is an in-sample
calibration outcome and not an independent estimate of screening recall. We
report it as such throughout, with a confidence interval, and
Section~\ref{sec:limitations} treats this as a primary limitation.

We also note what is missing: no result is reported at $\tau_{\text{ref}}$, and
we did not sweep $\tau$ to obtain a recall--specificity curve. The consequence
is quantified in Section~\ref{sec:forwarding_analysis}: because
$\tau = 0.00245$ yields a 49.3\% false-positive rate, the forwarding rate is
dominated by false alarms, and a calibrated operating point between $0.00245$
and $0.0189$ is the single most promising unexplored change to the system.

\subsection{Stage~2: YOLOv5n Defect Localisation}

Stage~2 is YOLOv5n \cite{jocher2022yolov5} (CSPDarknet backbone, SPPF, PANet)
with $320\times320$ RGB input, predicting bounding boxes and class
probabilities for the four defect types. We selected it over YOLOv5s and
YOLOv8n primarily on footprint: 1.76\,M parameters, 4.1\,GFLOPs, and 3.7\,MB,
against 14.3\,MB for YOLOv5s, on a device with 4\,GB of shared memory.
Section~\ref{sec:stage2_results} reports the comparison and is explicit that
the accuracy differences among the candidates are within the resolution of our
validation set; the choice rests on size, not on a demonstrated accuracy
advantage.

\subsection{Parallel Two-Stage Execution}
\label{sec:parallel_execution}

The deployed runtime is a producer--consumer pipeline. Images are enumerated
once per run. A loader thread performs OpenCV \texttt{imread} and pushes decoded
frames into a bounded queue (depth 8), while the main thread runs TensorRT
Stage~1 on each frame. When Stage~1 flags a frame, the same decoded
full-resolution BGR frame is passed through a multiprocessing queue
(\texttt{spawn} context, depth 8) to a dedicated child process running TensorRT
Stage~2 (confidence 0.25, IoU 0.45, letterbox $320\times320$). Stage~2 thus
avoids a second disk read. Overlap arises among JPEG decode, Stage~1 on the main
thread, and Stage~2 in the worker. End-to-end wall time starts before the
consumer loop and ends after the worker returns its accumulated Stage-2 time,
with the pipeline drained.

\paragraph{A scheduling asymmetry that we quantify rather than absorb.}
The YOLO-only and AE-only baselines use a single-threaded loop:
\texttt{imread} followed by inference, per image. They share the same image
folders and the same TensorRT engines as the parallel configuration, but they do
\emph{not} overlap decode with inference. The parallel configuration therefore
enjoys two advantages simultaneously---it invokes the detector less often, and
it hides decode latency---and a single end-to-end ratio cannot distinguish them.
Section~\ref{sec:where_the_speedup_goes} separates the two contributions
arithmetically from our own per-stage timings. We did not implement a threaded
YOLO-only control, and Section~\ref{sec:limitations} records this as the most
consequential missing experiment in the paper.

\paragraph{Timing model.}
End-to-end throughput is not the sum of per-frame Stage-1 and Stage-2 TensorRT
times; it is bounded by queueing, ingest, and cross-frame overlap. Let
$t_{\text{AE}}$ and $t_{\text{YOLO}}$ be mean TensorRT times for one pass
through each stage and let $p \in [0,1]$ be the \emph{forwarding rate}, the
fraction of frames Stage~1 sends onward. A first-order steady-state model for
wall time per input frame is
\begin{equation}
t_{\text{wall}} \approx \max\!\left(t_{\text{AE}},\; p \cdot t_{\text{YOLO}}\right) + t_{\text{overhead}},
\label{eq:twall}
\end{equation}
where $t_{\text{overhead}}$ captures JPEG decode, host copies, synchronisation,
and queue wait beyond the TensorRT kernels. Two caveats apply. First,
Eq.~\eqref{eq:twall} assumes the stages pipeline cleanly, whereas both contend
for the Nano's single 128-core GPU, so it is an optimistic bound on the
inference term. Second, we substitute measured values into
Eq.~\eqref{eq:twall} in Section~\ref{sec:where_the_speedup_goes} rather than
leaving it unevaluated, and the substitution is what reveals that
$t_{\text{overhead}}$ dominates both configurations. Speedup is reported
directly from measured throughputs as
$\mathrm{FPS}_{\text{parallel}}/\mathrm{FPS}_{\text{YOLO-only}}$.

\section{Experimental Setup}
\label{sec:setup}

\subsection{Dataset and Preprocessing}

We assembled 4,188 annotated images of industrial fabric, split into 3,939
training and 249 held-out images (94.0\%/6.0\%; Table~\ref{tab:data}).
Annotations are bounding boxes in YOLO format; per-class \emph{instance} counts
appear in Table~\ref{tab:classes}. Instance counts exceed image counts because a
single image may contain several defects. Sources are four Roboflow Universe
datasets, the ISL-Knit benchmark \cite{dasgupta2024islknit}, and a set of
AI-generated and in-house images collected in Bangladesh.

\begin{table}[H]
\caption{Dataset composition by source (4,188 images). Real and synthetic
in-house images are listed separately; no synthetic image is used for
evaluation.}
\label{tab:data}
\centering
\scriptsize
\setlength{\tabcolsep}{4pt}%
\begin{tabular}{@{}lrl@{}}
\hline
\textbf{Source} & \textbf{Images} & \textbf{Notes} \\
\hline
Fabric Detection \cite{roboflow-fabric-n3ili} & 89 & Roboflow (DENEME) \\
ISL-Knit \cite{dasgupta2024islknit} & 2,587 & Bangladesh knit dyeing \\
Fabric \cite{roboflow-fabric-5pxpq} & 76 & Roboflow (Afrasiab) \\
defect-needle \cite{roboflow-defect-needle-3pfnt} & 32 & Roboflow (ecovers) \\
Fabric Defect Augmented \cite{roboflow-fabric-defect-augmented-fdfmh} & 209 & Roboflow (TrainerWork) \\
Multifab (real) & 195 & Bangladesh garment factory \\
Multifab (Gemini, synthetic) & 1,000 & Training only \\
\hline
\textbf{Total} & \textbf{4,188} & \\
\hline
\end{tabular}
\end{table}

\begin{table}[H]
\caption{Defect class distribution (instance counts; 20,129 total)}
\label{tab:classes}
\centering
\scriptsize
\begin{tabular}{@{}lrr@{}}
\hline
\textbf{Class} & \textbf{Instances} & \textbf{Share} \\
\hline
Fly yarn     & 1,681  & 8.4\% \\
Hole         & 2,825  & 14.0\% \\
Oil spot     & 13,224 & 65.7\% \\
Thread error & 2,399  & 11.9\% \\
\hline
\textbf{Total} & \textbf{20,129} & 100\% \\
\hline
\end{tabular}
\end{table}

\paragraph{Synthetic data, and why it does not affect the reported metrics.}
Of the 1,195 in-house Multifab images, 195 are real production-line captures and
1,000 are synthetic augmentations generated with Google's Gemini model. For each
real defect image we prompted the model to (i)~relocate the defect within the
fabric, (ii)~produce similar defect patterns with realistic variation, and
(iii)~preserve fabric texture consistency; every sample was inspected visually
before inclusion. Synthetic images address class imbalance and the scarcity of
real defect examples.

Synthetic imagery can be easier both to reconstruct and to detect than real
fabric, which would inflate any metric computed over it. \textbf{We therefore
confine synthetic images to training: the 249-image benchmark contains only real
images, drawn from the ISL-Knit and Multifab (real) sources, and no
AI-generated image is used for evaluation anywhere in this paper.} All reported
Stage-1, Stage-2, and throughput figures therefore reflect performance on real
factory data.

\paragraph{Stage-1 training data.}
The autoencoder is trained on normal (defect-free) RGB patches only: images from
the training split carrying no annotated defect instance, approximately 800
images drawn primarily from ISL-Knit. Patches are resized to $120\times120$ and
normalised to $[0,1]$, with horizontal and vertical flips and ColorJitter
(brightness and contrast $\pm10\%$). The normal-only set used to compute the
reference threshold $\tau_{\text{ref}}$ (Section~\ref{sec:thresholding}) is a
held-out 10\% partition of these $\sim$800 images and is disjoint from the
autoencoder's training partition. Low Stage-1 resolution is deliberate: it
reduces edge compute, and $120\times120$ was sufficient for screening in our
tests.

\paragraph{Stage-2 training data.}
YOLOv5n is trained on the full annotated training split with letterboxing to
$320\times320$ and mosaic plus HSV augmentation.

\subsection{Training Configuration}

YOLOv5n (Ultralytics) was trained for 120 epochs, batch size 16, SGD with
initial learning rate 0.01 and weight decay $5\times10^{-4}$, on a cloud GPU
(Google Colab, Tesla T4). The autoencoder followed for 60 epochs, batch size 32,
Adam at $10^{-3}$ with ReduceLROnPlateau (factor 0.5, patience 5), using the
edge-weighted MSE of Eq.~\eqref{eq:edge_loss} and distillation at
$\lambda_{\text{kd}}=0.3$.

For deployment both models were exported to ONNX and then to TensorRT
\cite{nvidia2019tensorrt} FP16 on the Jetson Nano (JetPack 4.6, TensorRT 8.x).
FP32 YOLO inference exceeded available memory on the device, making FP16
conversion a hardware requirement rather than an optional optimisation. All
engines used in timed benchmarks are FP16.

\paragraph{FP16 and the Stage-1 threshold.}
Stage~1 compares a floating-point reconstruction error against a fixed
threshold, and FP16 conversion can perturb that error. We calibrated $\tau$ on
FP32 outputs and applied the same $\tau$ to the FP16 engine without
re-validating screening accuracy under FP16. Section~\ref{sec:limitations}
records this as an unquantified risk to the deployment claim.

\paragraph{Jetson measurement conditions.}
Measurements used an NVIDIA Jetson Nano Developer Kit on L4T R32.7.2 (kernel
4.9.253-tegra) in MAXN power mode, with OpenCV 4.12.0 for disk-based loading and
preprocessing. Jetson timings vary with thermal state; we report representative
runs after warm-up. We did not repeat runs enough times to report mean and
standard deviation, which Section~\ref{sec:limitations} notes.

\subsection{Evaluation Protocol}

\paragraph{Primary benchmark (249 images).}
Stage~1 screening and all Jetson timing use the same held-out 249-image set
(20 defective, 229 non-defective), disjoint from the 3,939-image training split.
Defect prevalence on this set is $20/249 = 8.0\%$. We report Stage-1 recall,
false positives, false-positive rate, specificity, and forwarding rate $p$, each
with an exact binomial (Clopper--Pearson) 95\% confidence interval. Three Jetson
configurations are timed on this set with identical engines: YOLO-only,
AE-only, and the parallel two-stage pipeline. All latencies are wall-clock, and
we report inference-only and end-to-end wall time separately.

\paragraph{Stage-2 architecture selection (202 images).}
The YOLOv5n / YOLOv5s / YOLOv8n comparison and the image-level accuracies in
Section~\ref{sec:stage2_results} were computed on a separate 202-image
validation split at $320\times320$, produced by a $95\%/5\%$ partition of the
4,188 images into 3,986 training and 202 validation. That split is
\textbf{defect-rich} --- 172 defective and 30 clean images --- because it was
built for detector selection, where clean images carry little signal.

\paragraph{The two evaluation sets are distinct and are not interchangeable.}
The 249-image benchmark (20 defective, 229 non-defective, 8.0\% prevalence) and
the 202-image split (172 defective, 30 clean, 85.1\% prevalence) differ in size,
in composition, and in purpose. We report each metric against the set that
produced it and never pool them. In particular, the Stage-1 and throughput
results in Sections~\ref{sec:stage1_results} and \ref{sec:jetson_results}
belong to the 249-image benchmark, and the Stage-2 architecture comparison in
Section~\ref{sec:stage2_results} belongs to the 202-image split. Because the two
sets have very different defect prevalence, an image-level accuracy computed on
one says nothing about accuracy on the other, and the forwarding-rate analysis
of Section~\ref{sec:forwarding_analysis} applies only to the 249-image
benchmark. Image-level results are reported as integer counts throughout so that
every percentage can be checked against its denominator.

\paragraph{Statistical reporting.}
Proportions carry Clopper--Pearson 95\% intervals. Comparisons between two
proportions on the same denominator use a two-proportion $z$-test, and we
report $p$-values rather than asserting differences. We flag explicitly where a
difference amounts to one or two images and is therefore not evidential.

\paragraph{Metrics we do not report, and why.}
We report no AUROC, no threshold sweep, and no per-class breakdown for either
stage. These are standard for anomaly detection and their absence is a genuine
limitation rather than an oversight of framing; Section~\ref{sec:limitations}
states what each would require.

\section{Results}
\label{sec:results}

\subsection{Stage~1: Anomaly Screening}
\label{sec:stage1_results}

\subsubsection{Canny Baseline}

We first tried a Canny-edge Stage~1 \cite{canny1986edge} as a zero-learning
reference. At a sensitive threshold (0.0009) it recovered 16/20 defects but
flagged almost every normal frame, leaving overall accuracy at 10.4\%. Relaxing
the threshold to 0.12 raised accuracy to 55.0\% but dropped defect recall to
7/20. No single threshold gave acceptable recall and a usable false-positive
rate simultaneously (Table~\ref{tab:canny}), which motivated a learned screener.

\begin{table}[H]
\caption{Canny Stage~1 on the 249-image benchmark. Neither threshold is usable:
the low setting flags nearly all normal frames, the high setting misses most
defects.}
\label{tab:canny}
\centering
\scriptsize
\setlength{\tabcolsep}{3pt}%
\begin{tabular}{@{}lcc@{}}
\hline
Threshold & Recall (defects) & Overall accuracy \\
\hline
0.0009 (low)  & 80\% (16/20) & 10.4\% \\
0.12 (high)   & 35\% (7/20)  & 55.0\% \\
\hline
\end{tabular}
\end{table}

\noindent Canny is an edge detector, not an anomaly detector, so this
establishes only that hand-crafted edge energy is insufficient here. It is not a
competitive baseline, and we do not present it as one
(Section~\ref{sec:limitations}).

\subsubsection{Base vs.\ Modified Autoencoder}

Table~\ref{tab:ae} compares the base autoencoder (plain MSE) against the
modified autoencoder (attention gates, edge-weighted loss, and distillation
together) at $\tau = 0.00245$. We report absolute counts, rates, and intervals,
because the relative improvement alone is misleading about the operating point.

\begin{table}[H]
\caption{Base vs.\ modified autoencoder on the 249-image benchmark (20
defective, 229 non-defective) at $\tau=0.00245$. FPR = false-positive rate on
the 229 non-defective images. Brackets are Clopper--Pearson 95\% intervals. The
final row gives the trivial classifier that labels every frame normal, included
because the accuracy column is otherwise easy to misread.}
\label{tab:ae}
\centering
\scriptsize
\setlength{\tabcolsep}{3pt}%
\begin{tabular}{@{}lcccc@{}}
\hline
Model & Recall & FP & FPR & Acc. \\
\hline
Base AE & 19/20 & 140/229 & 0.611 & 0.434 \\
        & \scriptsize[0.75, 1.00] & & \scriptsize[0.54, 0.67] & \\
Modified AE & \textbf{20/20} & \textbf{113/229} & \textbf{0.493} & 0.546 \\
        & \scriptsize[0.83, 1.00] & & \scriptsize[0.43, 0.56] & \\
\hline
\emph{Always-normal} & 0/20 & 0/229 & 0.000 & \emph{0.920} \\
\hline
\end{tabular}
\end{table}

\paragraph{What the numbers support.}
The reduction in false positives from 140 to 113 out of 229 is a 19.3\%
relative reduction and is statistically supported ($z = 2.56$, $p = 0.011$,
two-proportion test). This is the one Stage-1 comparison in this paper that
survives a significance test, and it is why we adopted the modified autoencoder:
fewer spurious forwards means less Stage-2 load on the Jetson.

\paragraph{What the numbers do not support.}
Three cautions apply, and we state them here rather than in a closing paragraph.

First, the recall improvement from 19/20 to 20/20 is a \textbf{single image}. The
two intervals, $[0.75, 1.00]$ and $[0.83, 1.00]$, overlap almost completely. This
difference is not evidence for the modified architecture and we do not treat it
as such.

Second, \textbf{20/20 is an in-sample calibration outcome}. As
Section~\ref{sec:thresholding} states, $\tau = 0.00245$ was chosen on this
benchmark to prioritise recall. Even taken at face value, 20/20 on $n=20$ has a
95\% interval of $[0.83, 1.00]$: these data cannot exclude a true screening
recall as low as 83\%, which for a system whose safety argument rests on high
recall is the most important number in this subsection.

Third, \textbf{the accuracy column is below the majority-class baseline}. On a
set that is 92.0\% non-defective, labelling everything normal scores 0.920,
against 0.546 for the modified autoencoder. Accuracy is the wrong summary for
this class balance; we print it only for continuity with the recall/FPR pair and
include the trivial classifier so the comparison is unavoidable. The operative
figure is that Stage~1 forwards \textbf{49.3\%} of clean fabric --- it rejects
about half of normal frames, not the "vast majority" that a cascade
motivation implicitly assumes. Section~\ref{sec:forwarding_analysis} works
through what this costs.

\paragraph{Ablation not performed.}
The modified model bundles three changes. Table~\ref{tab:ae} cannot attribute
the 19.3\% reduction to attention gating, edge weighting, or distillation
individually, and we make no such attribution anywhere in this paper.

\begin{figure}[H]
\centering
\begin{tikzpicture}
\begin{axis}[
  ybar,
  width=0.7\columnwidth,
  height=5.2cm,
  bar width=10pt,
  symbolic x coords={Defects found, False pos., Normals passed},
  xtick=data,
  ylabel=Count,
  ymin=0, ymax=150,
  legend style={at={(0.5,-0.2)},anchor=north,legend columns=2}]
\addplot coordinates {(Defects found,19) (False pos.,140) (Normals passed,89)};
\addplot coordinates {(Defects found,20) (False pos.,113) (Normals passed,116)};
\legend{Base AE, Modified AE}
\end{axis}
\end{tikzpicture}
\caption{Base vs.\ modified autoencoder: defects found (of 20), false positives,
and normal images correctly passed (of 229). The false-positive reduction is the
supported effect; the one-image recall difference is not.}
\label{fig:ae_base_vs_modified}
\end{figure}
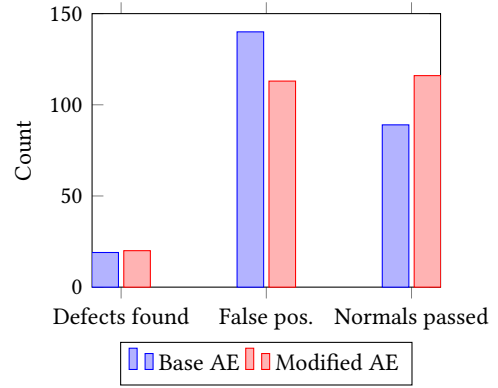

\subsection{Stage~2: Defect Detection}
\label{sec:stage2_results}

Table~\ref{tab:yolo} reports detection metrics for the three Stage-2 candidates
and Table~\ref{tab:yolo_image} their image-level accuracy on the 202-image
validation split.

\begin{table}[H]
\caption{Stage-2 candidates at $320\times320$. Box-level metrics on the
validation split.}
\label{tab:yolo}
\centering
\scriptsize
\setlength{\tabcolsep}{2pt}%
\begin{tabular}{@{}lrrrrrrr@{}}
\hline
Model & Par. & GFLOPs & mAP50 & mAP50-95 & Prec. & Rec. & Size \\
\hline
YOLOv5n & 1.76M & 4.1  & 0.742 & 0.363 & 0.825 & 0.637 & 3.7 MB \\
YOLOv5s & 7.02M & 15.8 & 0.716 & 0.345 & 0.775 & 0.653 & 14.3 MB \\
YOLOv8n & 3.01M & 8.1  & 0.676 & 0.339 & 0.759 & 0.604 & 6.2 MB \\
\hline
\end{tabular}
\end{table}

\begin{table}[H]
\caption{Image-level performance on the 202-image validation split (172
defective, 30 clean), with confusion-matrix counts and Clopper--Pearson 95\%
intervals on accuracy. YOLOv5n and YOLOv5s differ by two images; YOLOv8n is
separated from both by its 20 false negatives.}
\label{tab:yolo_image}
\centering
\scriptsize
\setlength{\tabcolsep}{2.5pt}%
\begin{tabular}{@{}lrrrrcc@{}}
\hline
Model & TN & FP & FN & TP & Acc. & 95\% CI \\
\hline
YOLOv5n & 28 & 2 & 1  & 171 & 199/202 = 98.5\% & [0.957, 0.997] \\
YOLOv5s & 29 & 1 & 0  & 172 & 201/202 = 99.5\% & [0.973, 1.000] \\
YOLOv8n & 30 & 0 & 20 & 152 & 182/202 = 90.1\% & [0.851, 0.939] \\
\hline
\end{tabular}
\end{table}

\paragraph{Model selection rests on size, not accuracy.}
YOLOv5n has the highest mAP@0.5 (0.742) and the smallest footprint, but the
accuracy difference against YOLOv5s is not resolvable at this sample size: the
two differ by \textbf{two images} out of 202, with overlapping intervals, and
YOLOv5s is in fact the marginally more accurate of the pair (201/202 against
199/202). YOLOv8n is separable from both, and the confusion matrix identifies
why: its 20 false negatives against 1 for YOLOv5n mean it misses defective
images outright rather than mislabelling clean ones, which is the failure mode
least acceptable in a screening cascade.

We therefore select YOLOv5n on the grounds that survive scrutiny --- 3.7\,MB and
4.1\,GFLOPs against 14.3\,MB and 15.8\,GFLOPs for YOLOv5s, on a 4\,GB device ---
and explicitly \emph{not} on an accuracy advantage over YOLOv5s, which the data
do not support. Earlier drafts of this work described YOLOv5n as best on
accuracy; that claim is withdrawn.

\paragraph{Selection and reporting use the same split.}
The three candidates were compared on this split and the selected model's
accuracy is reported on it. There is no separate test set. The reported accuracy
is therefore optimistically biased by selection over three architectures, and
the mAP@0.5 figures are best-epoch validation values rather than values from a
held-out set.

\paragraph{Localisation is weak, and image-level accuracy hides it.}
Box-level recall for YOLOv5n is 0.637, so roughly one defect instance in three
is missed at the localisation level, and mAP@0.5:0.95 is 0.363. Yet image-level
accuracy is 98.5\%. Both statements are true: most defective \emph{images}
contain at least one detected instance even when other instances in the same
image are missed. For triage --- flagging a roll region for human review ---
image-level behaviour is the operative property, which is why we report it. But
this system should not be described as accurately localising or counting
defects, and any downstream use requiring per-instance counts is out of scope.
The likely causes are small defect size and the $320\times320$ input; higher
resolution or test-time augmentation are natural next steps.

\paragraph{Class imbalance.}
Oil spot accounts for 13,224 of 20,129 annotated instances (65.7\%) while fly
yarn accounts for 1,681 (8.4\%). The aggregate mAP@0.5 of 0.742 may therefore be
driven disproportionately by oil spot, and we report no per-class metrics
(Section~\ref{sec:limitations}). Aggregate mAP on this distribution should not
be read as evidence of performance on minority defect types.

\begin{figure}[H]
\centering
\includegraphics[width=\columnwidth]{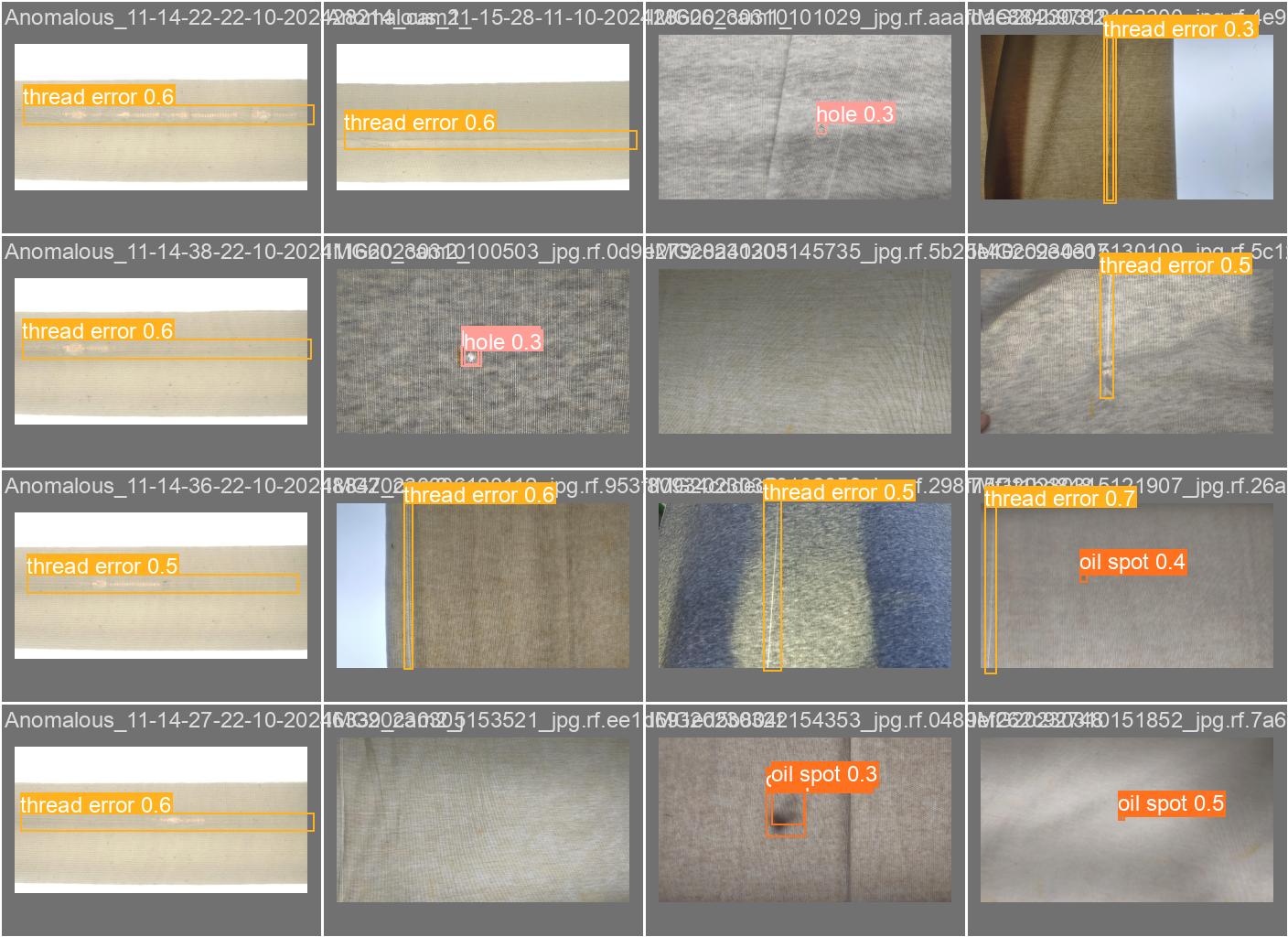}
\caption{\textbf{YOLOv5n detection example} ($320\times320$): bounding boxes and
confidence scores for the four defect classes.}
\label{fig:yolo_detection}
\end{figure}

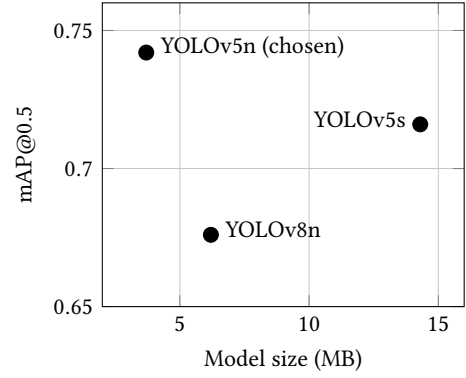
\begin{figure}[H]
\centering
\begin{tikzpicture}
\begin{axis}[
  width=0.72\columnwidth,
  height=5.6cm,
  xlabel={Model size (MB)},
  ylabel={mAP@0.5},
  xmin=2, xmax=16,
  ymin=0.65, ymax=0.76,
  grid=both,
  major grid style={line width=.2pt,draw=gray!45},
  minor grid style={line width=.1pt,draw=gray!25},
]
\addplot[only marks, mark=*, mark size=2.8pt] coordinates {
  (3.7,0.742) (14.3,0.716) (6.2,0.676)
};
\node[anchor=west,font=\footnotesize] at (axis cs:3.9,0.744) {YOLOv5n (chosen)};
\node[anchor=east,font=\footnotesize] at (axis cs:14.1,0.718) {YOLOv5s};
\node[anchor=west,font=\footnotesize] at (axis cs:6.4,0.678) {YOLOv8n};
\end{axis}
\end{tikzpicture}
\caption{Model size versus mAP@0.5 for the Stage-2 candidates. YOLOv5n is
selected for footprint; the mAP differences among the three are within the
resolution of the validation split.}
\label{fig:yolo_size_map_tradeoff}
\end{figure}

\FloatBarrier
\subsection{Jetson Nano Throughput}
\label{sec:jetson_results}

Table~\ref{tab:jetson_e2e} reports the three configurations on the 249-image
benchmark with identical FP16 engines. Inference-only time is measured inside
TensorRT \texttt{execute()}; end-to-end wall time is total elapsed time divided
by $N$, including JPEG decode, in-detector preprocessing, multiprocessing
handoff, and synchronisation.

\begin{table}[H]
\caption{Jetson Nano runtime on the 249-image benchmark (TensorRT FP16,
$\tau=0.00245$, $p=0.534$). Inference for the two-stage row is
$t_{\mathrm{AE}} + p\,t_{\mathrm{YOLO}} = 22.67 + 0.534(42.47)$. The overhead
column is wall minus inference and is the quantity that turns out to matter.}
\label{tab:jetson_e2e}
\centering
\scriptsize
\setlength{\tabcolsep}{3pt}%
\begin{tabular}{@{}lrrrr@{}}
\hline
\textbf{Configuration} & \textbf{Inf.} & \textbf{Wall} & \textbf{Overhead} & \textbf{FPS} \\
 & (ms) & (ms) & (ms) & \\
\hline
YOLO-only (sequential) & 47.77 & 101.40 & 53.63 & 9.86 \\
AE-only (sequential)   & 27.29 & 82.25  & 54.96 & 12.16 \\
Two-stage (parallel)   & 45.36 & 74.33  & \textbf{28.97} & \textbf{13.45} \\
\hline
\end{tabular}
\end{table}

The parallel pipeline reaches 13.45\,FPS against 9.86\,FPS for sequential
YOLO-only, a ratio of $1.36\times$. Reported alone, that ratio would be
misleading, and the next subsection explains why.

\subsection{Where the Speedup Actually Comes From}
\label{sec:where_the_speedup_goes}

The parallel configuration differs from the baselines in \emph{two} ways at
once: it invokes the detector on only 53.4\% of frames, and it overlaps JPEG
decode with inference (Section~\ref{sec:parallel_execution}). A single
end-to-end ratio cannot separate these. Our own per-stage timings can.

Total wall-time gain over YOLO-only is $101.40 - 74.33 = 27.07$\,ms per image.
Decomposing it:
\begin{align}
\Delta_{\text{inference}} &= 47.77 - 45.36 = 2.41~\text{ms} \quad (8.9\%), \label{eq:dinf}\\
\Delta_{\text{overhead}}  &= 53.63 - 28.97 = 24.66~\text{ms} \quad (91.1\%). \label{eq:dovh}
\end{align}

\noindent \textbf{The cascade contributes 8.9\% of the measured gain; overlapping
the data path contributes 91.1\%.} At $p = 0.534$ the cascade reduces inference
time by only 5.1\% (45.36 against 47.77\,ms), because Stage~2 still runs on more
than half of all frames while Stage~1 now runs on all of them.

The reason is visible in the overhead column of Table~\ref{tab:jetson_e2e}.
Non-inference cost is 53.63\,ms per image for sequential YOLO-only and 54.96\,ms
for sequential AE-only --- consistent, as expected, since both decode the same
JPEGs --- and \emph{exceeds} the 47.77\,ms spent inside TensorRT. On this
hardware the pipeline is ingest-bound, not inference-bound. A cascade cannot
save what is not being spent on the detector.

\paragraph{Substituting into the timing model.}
Eq.~\eqref{eq:twall} makes the same point. For the two-stage configuration,
$\max(t_{\text{AE}},\, p\,t_{\text{YOLO}}) = \max(22.67,\, 22.68) = 22.68$\,ms
against a measured 74.33\,ms, giving $t_{\text{overhead}} = 51.65$\,ms. For
YOLO-only, $\max(0,\, 1.0 \times 47.77) = 47.77$ against 101.40, giving
$t_{\text{overhead}} = 53.63$\,ms. The two estimates agree to within 2\,ms and
both put overhead above 51\,ms --- larger than either configuration's inference
term. The model we wrote down predicts that this system is dominated by
$t_{\text{overhead}}$, and the measurements confirm it.

\paragraph{What a fair baseline would likely show.}
We did not implement a threaded YOLO-only control, so the following is an
estimate rather than a measurement. If YOLO-only used the same loader thread and
its decode were hidden behind inference, Eq.~\eqref{eq:twall} gives
$t_{\text{wall}} \approx \max(53.63,\, 47.77) = 53.63$\,ms, or
\textbf{18.65\,FPS} --- above the 13.45\,FPS of the two-stage pipeline. On that
estimate the $1.36\times$ advantage would not merely shrink, it would
\textbf{invert to roughly $0.72\times$}, because the two-stage configuration runs
Stage~1 on every frame in addition to Stage~2 on half of them.

We therefore do not claim that cascading accelerates this system. What we claim
is narrower and better supported: the cascade reduces detector invocations by
46.6\% and inference time by 5.1\% at the measured operating point, and the
observed throughput gain over our sequential baselines is predominantly an
artefact of pipelining the data path --- an optimisation available to any
configuration, including YOLO-only.

\paragraph{Implication for the literature.}
Cascade speedups measured against a sequential baseline conflate two effects
whose magnitudes here differ by an order of magnitude. We suggest that
end-to-end cascade results be reported with the ingest path controlled, or with
the decomposition of Eqs.~\eqref{eq:dinf}--\eqref{eq:dovh} made explicit. On a
live camera feed, where \texttt{imread} is replaced by capture and buffering,
$t_{\text{overhead}}$ would change and so would this balance; the direction of
that change is not something our disk-based benchmark can establish.

\begin{figure}[H]
\centering
\begin{tikzpicture}
\begin{axis}[
  ybar stacked,
  width=0.82\columnwidth,
  height=5.4cm,
  bar width=16pt,
  enlarge x limits=0.3,
  x tick label style={rotate=15, anchor=east},
  symbolic x coords={YOLO-only, Two-stage},
  xtick=data,
  ylabel={ms per image},
  ymin=0, ymax=110,
  legend style={at={(0.5,-0.28)},anchor=north,legend columns=2}]
\addplot coordinates {(YOLO-only,47.77) (Two-stage,45.36)};
\addplot coordinates {(YOLO-only,53.63) (Two-stage,28.97)};
\legend{Inference, Non-inference (decode, copies, sync)}
\end{axis}
\end{tikzpicture}
\caption{Wall time decomposed. The inference bars are nearly equal (47.77 vs.\
45.36\,ms): the cascade saves little detector compute at $p=0.534$. Almost the
entire difference in total height is non-inference time, which the parallel
runtime overlaps and the sequential baseline does not.}
\label{fig:jetson_decomposition}
\end{figure}
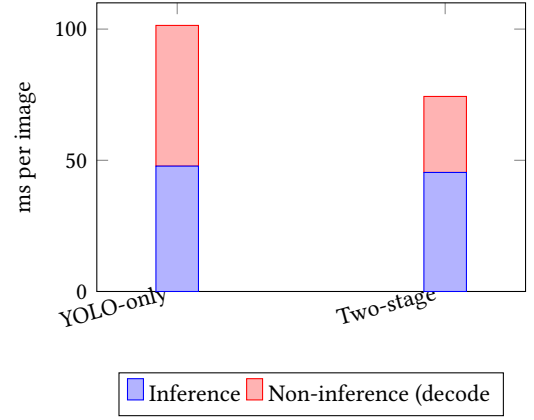

\subsection{What Limits the Forwarding Rate}
\label{sec:forwarding_analysis}

The forwarding rate is not a property of the screener alone. Writing $\pi$ for
defect prevalence, $R$ for Stage-1 recall, and $F$ for its false-positive rate,
\begin{equation}
p = \pi R + (1-\pi) F .
\label{eq:fwd}
\end{equation}
Substituting our measured values ($\pi = 20/249 = 0.0803$, $R = 1.00$,
$F = 113/229 = 0.4934$) gives $p = 0.0803 + 0.9197 \times 0.4934 = 0.534$,
reproducing the measured forwarding rate exactly and confirming that
Eq.~\eqref{eq:fwd} accounts for it.

\paragraph{Forwarding here is false-positive-limited.}
Splitting the two terms is informative:
\[
\underbrace{0.0803}_{\text{true defects, }15.0\%} \;+\;
\underbrace{0.4538}_{\text{false positives, }85.0\%} \;=\; 0.534 .
\]
\textbf{85\% of forwarded frames are false alarms.} The consequence is that
lowering prevalence barely helps: holding $\tau$ fixed, $p$ falls only from
0.534 at $\pi=0.080$ to 0.499 at $\pi=0.01$, because the $(1-\pi)F$ term
dominates. The cascade's compute saving on this benchmark is therefore bounded
by Stage-1 \emph{specificity}, not by how rare defects are --- the opposite of
what the sparsity motivation would suggest, and a distinction that a single
reported $p$ conceals.

\paragraph{What tighter calibration would buy.}
Table~\ref{tab:fwd_projection} evaluates Eq.~\eqref{eq:fwd} at
$\pi = 0.0803$ across hypothetical false-positive rates. These rows are
\emph{projections}, not measurements: they assume recall is preserved as $\tau$
rises, which we did not verify and which the base autoencoder's behaviour gives
no reason to take for granted.

\begin{table}[H]
\caption{Projected forwarding rate and inference cost at $\pi=0.0803$ as a
function of Stage-1 false-positive rate, from Eq.~\eqref{eq:fwd} with
$t_{\mathrm{AE}}=22.67$\,ms and $t_{\mathrm{YOLO}}=42.47$\,ms. Only the first
row is measured; the rest assume recall is preserved and are upper bounds on the
achievable benefit.}
\label{tab:fwd_projection}
\centering
\scriptsize
\setlength{\tabcolsep}{3pt}%
\begin{tabular}{@{}lrrrr@{}}
\hline
 & \textbf{FPR} & \textbf{$p$} & \textbf{Inf.\ (ms)} & \textbf{vs.\ YOLO} \\
\hline
Measured ($\tau{=}0.00245$) & 0.493 & 0.534 & 45.35 & $-5.1\%$ \\
Projected & 0.200 & 0.264 & 33.89 & $-29.0\%$ \\
Projected & 0.100 & 0.172 & 29.99 & $-37.2\%$ \\
Projected & 0.050 & 0.126 & 28.03 & $-41.3\%$ \\
Projected & 0.010 & 0.090 & 26.47 & $-44.6\%$ \\
\hline
\end{tabular}
\end{table}

The reference threshold of Section~\ref{sec:thresholding},
$\tau_{\text{ref}} \approx 0.0189$, targets roughly 1\% false positives by
construction --- the last row of Table~\ref{tab:fwd_projection}, a projected
44.6\% inference reduction against the measured 5.1\%. Whether recall survives
at that threshold is precisely the experiment we did not run, and it is the
single highest-value addition to this work. Note also that even a 44.6\%
inference reduction would not by itself yield a large throughput gain on this
hardware, because overhead would still exceed inference
(Section~\ref{sec:where_the_speedup_goes}); reducing $p$ and pipelining the
ingest path are complementary, and neither alone is sufficient.

\subsection{Discussion}

Three findings, in descending order of how well we can support them.

\textbf{A learned screener is necessary.} No Canny threshold achieves usable
recall and specificity together (Table~\ref{tab:canny}). This is the most
robust claim in the paper, though it establishes only a lower bar.

\textbf{The modified autoencoder reduces false forwards.} 140 to 113 of 229,
$p=0.011$ (Table~\ref{tab:ae}). We cannot attribute this to any individual
component, and the accompanying recall difference of one image is not
evidential.

\textbf{The cascade's throughput gain is mostly not the cascade.} Of the
measured $1.36\times$, 91.1\% traces to overlapping the data path and 8.9\% to
reduced detector invocation (Eqs.~\eqref{eq:dinf}--\eqref{eq:dovh}). Against a
comparably threaded YOLO-only baseline we estimate the advantage would invert.
We consider this the most useful result here, because it is the one most likely
to change how a practitioner reads cascade throughput numbers --- and because it
identifies the actual bottleneck on this class of device as image ingest.

Taken together these point to a specific engineering conclusion: on a Jetson
Nano running from disk, the ordering of work that matters is (i) pipeline the
ingest path, which is worth $\sim$25\,ms per image, then (ii) recalibrate
Stage~1 to cut $p$, which Table~\ref{tab:fwd_projection} suggests is worth up to
$\sim$19\,ms of inference, and only then (iii) revisit the screener
architecture, whose measured contribution here is 27 false positives out of 229.
Our results order these three, and that ordering is the practical contribution
we would want carried forward.

\section{Limitations}
\label{sec:limitations}

We state limitations explicitly and in one place, ordered by how much they
constrain the conclusions.

\paragraph{L1. The Stage-1 threshold was selected on the evaluation set.}
$\tau = 0.00245$ was chosen on the 249-image benchmark to obtain full defect
recall, and recall is then reported on that same set
(Section~\ref{sec:thresholding}). The 20/20 figure is therefore an in-sample
calibration outcome, not an independent estimate. Reporting it with its
Clopper--Pearson interval $[0.83, 1.00]$ is the most we can honestly do with
these data; a genuine estimate requires a calibration split disjoint from the
evaluation set.

\paragraph{L2. Twenty defective images is too few for a recall claim.}
Every Stage-1 recall figure rests on $n=20$. The interval on 20/20 is
$[0.83, 1.00]$, so a true recall of 85\% is entirely consistent with our
observation. For a system whose safety case is "misses nothing", this is the
binding evidential weakness. It cannot be fixed by analysis, only by more
defective images.

\paragraph{L3. No threaded YOLO-only control.}
Our baselines are single-threaded while the two-stage configuration is
pipelined, so the measured $1.36\times$ conflates cascade savings with decode
overlap. Section~\ref{sec:where_the_speedup_goes} separates them
arithmetically and estimates that a threaded YOLO-only baseline would reach
$\sim$18.65\,FPS, inverting the comparison. That estimate is derived from
Eq.~\eqref{eq:twall} and our measured overhead, not measured directly. This is
the most consequential missing experiment in the paper and requires no
retraining --- only re-running the baseline with the loader thread enabled.

\paragraph{L4. No modern anomaly-detection baseline.}
Our only Stage-1 comparison points are Canny and a plain autoencoder. Canny is
an edge detector, not an anomaly detector, so it establishes a low bar rather
than a competitive one. We do not compare against PaDiM
\cite{defard2021padim}, PatchCore \cite{roth2022patchcore}, FastFlow
\cite{yu2021fastflow}, RD4AD \cite{deng2022rd4ad}, SimpleNet
\cite{liu2023simplenet}, or EfficientAD \cite{batzner2024efficientad}. Several
of these require no training on our data beyond fitting statistics over a frozen
ImageNet backbone, so their absence reflects scope rather than difficulty. No
claim in this paper should be read as asserting that our screener is competitive
with the anomaly-detection state of the art.

\paragraph{L5. No threshold sweep, no ROC, no AUROC.}
We report results at a single $\tau$ and describe a second value we never
evaluated. For a reconstruction-error anomaly detector, the recall--specificity
curve is the standard characterisation and its absence is why the operating
point in Section~\ref{sec:forwarding_analysis} remains
false-positive-limited without our being able to say what the achievable
frontier is. Table~\ref{tab:fwd_projection} is a projection under an unverified
recall assumption, not a measurement.

\paragraph{L6. No per-class metrics.}
Oil spot is 65.7\% of annotated instances and fly yarn 8.4\%
(Table~\ref{tab:classes}), so the aggregate mAP@0.5 of 0.742 is likely dominated
by the majority class, and Stage-1 recall is reported only in aggregate over 20
defective images. Performance on minority defect types is unmeasured.

\paragraph{L7. Components are not ablated.}
The modified autoencoder bundles attention gates, edge-weighted loss, and
distillation. Table~\ref{tab:ae} measures the bundle. Which component drives the
19.3\% false-positive reduction is unknown, and Section~\ref{sec:methodology}
notes a mechanism by which edge weighting could plausibly hurt rather than help.

\paragraph{L8. FP16 conversion is not accuracy-validated.}
Stage~1 thresholds a floating-point reconstruction error. We calibrated $\tau$ on
FP32 outputs and deployed FP16 engines without re-measuring screening accuracy
under FP16 (Section~\ref{sec:setup}). All accuracy results are therefore FP32
and all timing results FP16, and we cannot rule out that quantisation shifts the
error distribution relative to $\tau$. This is an unquantified risk to the
deployment claim specifically.

\paragraph{L9. No separate test set for Stage 2.}
Three architectures were compared on the 202-image validation split and the
selected model's accuracy is reported on the same split
(Section~\ref{sec:stage2_results}), so that figure is optimistically biased by
selection. The mAP values are best-epoch validation numbers.

\paragraph{L10. Single-run timings.}
Jetson measurements are representative post-warm-up runs, not means over
repetitions, and thermal state affects them. We report no variance, so small
differences between configurations should not be over-read.

\paragraph{L11. Disk-based benchmark, not a live feed.}
Frames are read from disk with \texttt{imread}. A camera stream would replace
decode with capture and buffering, changing $t_{\text{overhead}}$ and therefore
the balance in Section~\ref{sec:where_the_speedup_goes}. Since that overhead is
what dominates our measurements, the throughput numbers here should not be
extrapolated to a live deployment.

\paragraph{L12. Scope of evaluation.}
Four defect types, one fabric category, patch- and image-level metrics rather
than full-roll streaming metrics, and no continuous long-duration factory trial.
Synthetic images are 24\% of the corpus but are confined to training
(Section~\ref{sec:setup}), so the reported rates are measured on real fabric;
what remains untested is whether training on partly synthetic data biases the
learned notion of normality in ways a real-only training set would not.

\section{Conclusion}
\label{sec:conclusion}

We built a two-stage fabric defect cascade --- a compact autoencoder screener
followed by YOLOv5n on flagged frames --- and deployed it end to end on an
NVIDIA Jetson Nano with TensorRT FP16 engines and a concurrent
producer--consumer runtime. The system works: at a recall-prioritised threshold
Stage~1 flags all 20 defective images in our 249-image benchmark, the modified
autoencoder cuts false forwards from 140 to 113 of 229 non-defective images
($p=0.011$) relative to a plain autoencoder, and the parallel pipeline sustains
13.45\,FPS against 9.86\,FPS for a sequential YOLO-only loop.

The more useful contribution is what happens when that $1.36\times$ is taken
apart. Decomposing it into inference and non-inference components shows that
\textbf{91\% of the gain comes from overlapping JPEG decode with inference and
only 8.9\% from cascading}: at the measured forwarding rate $p = 0.534$ the
cascade reduces inference time by 5.1\%, because non-inference cost
(53.63\,ms/image) exceeds time spent inside TensorRT (47.77\,ms/image). On this
class of device the pipeline is ingest-bound, and against a comparably threaded
YOLO-only baseline we estimate the advantage would invert to roughly
$0.72\times$. We therefore do not claim that cascading accelerates this system,
and we suggest that cascade throughput results be reported with the ingest path
controlled or with this decomposition made explicit.

A second analysis explains why the cascade saves so little here. Decomposing
the forwarding rate as $p = \pi R + (1-\pi)F$ reproduces the measured
$p = 0.534$ exactly and shows that \textbf{85\% of forwarded frames are false
alarms}. Forwarding on this benchmark is limited by Stage-1 specificity rather
than by defect prevalence, so lowering prevalence barely reduces $p$ while
tightening calibration could reduce inference cost by a projected 29--45\%.
That inverts the usual reading of the sparsity argument: sparsity creates the
opportunity, but specificity determines whether it can be taken.

These two analyses order the engineering work for anyone deploying such a
system on comparable hardware: pipeline the ingest path first, recalibrate the
screening threshold second, and revisit the screener architecture last.

Our conclusions are constrained by the limitations in
Section~\ref{sec:limitations}, of which three matter most: the Stage-1
threshold was selected on the evaluation set, so 20/20 is an in-sample
calibration outcome with a 95\% interval of $[0.83, 1.00]$; there are only 20
defective images, which is too few to establish a recall claim; and we did not
run the threaded YOLO-only control that would settle the throughput comparison
directly. That control requires no retraining and is the first experiment we
would add. Beyond it, the highest-value next steps are a Stage-1 threshold
sweep with a disjoint calibration split, comparison against a modern anomaly
detector, per-class metrics, and validation of screening accuracy under FP16.

In its present form the system is AI-assisted screening and triage for
low-cost factories, not autonomous acceptance testing. What we can say with
confidence is narrower than a headline speedup but more portable: a defect
cascade fits on a \$100-class device, and on that device the cost that dominates
is getting pixels to the model rather than running the model.

\section*{Data and Code Availability}
The annotated dataset used in this work is published on Roboflow Universe as
\texttt{dfab} (version~7, four classes, Public Domain licence):
\url{https://universe.roboflow.com/personal-lczvh/dfab/dataset/7}. It
incorporates the ISL-Knit benchmark \cite{dasgupta2024islknit} and four Roboflow
Universe subsets under their respective licences, together with the Multifab
images described in Section~\ref{sec:setup}.

Training and evaluation code, the exported ONNX models, the TensorRT conversion
and Jetson benchmark scripts, and the per-image score files behind every table
are available from the corresponding author on reasonable request; we intend to
release them in a public repository with an archived DOI.

\section*{Acknowledgements}
The authors declare no competing interests.

\bibliographystyle{unsrt}
\bibliography{references}

\end{document}